 \documentclass[final,5p,times,twocolumn,authoryear]{elsarticle}

\usepackage{amssymb}
\usepackage{lipsum}
\usepackage{hyperref}
\usepackage{amsmath}
\usepackage{float}

\journal{arXiv}

\begin{document}

\begin{frontmatter}



\title{Multimodal Feature-Driven Deep Learning for the Prediction of Duck Body Dimensions and Weight}



\author[HUNAU,UNSW]{Wenbo Xiao\fnref{equal}}
\author[HUNAU]{Qiannan Han\fnref{equal}}
\author[SICAU]{Gang Shu}
\author[HUNAU]{Guiping Liang}
\author[HUNAU]{Hongyan Zhang}
\author[HUNAU]{Song Wang}
\author[HUNAU]{Zhihao Xu}
\author[Vet]{Weican Wan}
\author[Vet]{Chuang Li}

\author[Vet]{Guitao Jiang\corref{corresponding}}
\ead{jiangguitao@163.com}

\author[HUNAU]{Yi Xiao\corref{corresponding}}
\ead{xiaoyi@hunau.edu.cn}

\cortext[corresponding]{Corresponding author.}
\fntext[equal]{The two authors contribute equally to this work.}

\affiliation[HUNAU]{
organization={College of Information and Technology, Hunan Agricultural University},
city={Changsha},
postcode={410129},
state={Hunan},
country={China}
}
\affiliation[Vet]{
organization={Hunan Institute of Animal and Veterinary Science},
city={Changsha},
postcode={410131},
state={Hunan},
country={China}
}

\affiliation[SICAU]{
organization={College of Animal Science and Technology, Sichuan Agricultural University},
city={Chengdu},
postcode={611130},
state={Sichuan},
country={China}
}

\affiliation[UNSW]{
organization={School of Computer Science and Engineering, the University of New South Wales},
city={Sydney},
postcode={2052},
state={NSW},
country={Australia}
}

\begin{abstract}
Accurate body dimension and weight measurements are critical for optimizing poultry management, health assessment, and economic efficiency. This study introduces an innovative deep learning-based model leveraging multimodal data—2D RGB images from different views, depth images, and 3D point clouds—for the non-invasive estimation of duck body dimensions and weight. A dataset of 1,023 Linwu ducks, comprising over 5,000 samples with diverse postures and conditions, was collected to support model training. The proposed method innovatively employs PointNet++ to extract key feature points from point clouds, extracts and computes corresponding 3D geometric features, and fuses them with multi-view convolutional 2D features. A Transformer encoder is then utilized to capture long-range dependencies and refine feature interactions, thereby enhancing prediction robustness. The model achieved a mean absolute percentage error (MAPE) of 6.33\% and an R\textsuperscript{2} of 0.953 across eight morphometric parameters, demonstrating strong predictive capability. Unlike conventional manual measurements, the proposed model enables high-precision estimation while eliminating the necessity for physical handling, thereby reducing animal stress and broadening its application scope. This study marks the first application of deep learning techniques to poultry body dimension and weight estimation, providing a valuable reference for the intelligent and precise management of the livestock industry with far-reaching practical significance.

\end{abstract}




\end{frontmatter}




\section{Introduction}
\label{introduction}

Duck body measurements, including indices such as body diagonal length, keel length, chest depth, and chest width, along with weight, play a vital role in real-time monitoring of growth and development. These metrics are essential for understanding individual differences, identifying outliers (e.g., excessively large or small ducks), and diagnosing potential nutritional issues. By facilitating timely human intervention in feeding strategies, such measurements contribute to improving breeding efficiency and provide data-driven support for intelligent and precision farming.

Traditionally, duck body measurements rely on manual methods involving tape measures and calipers, which require additional human assistance to hold ducks stationary. This process is not only labor-intensive and time-consuming but also induces stress in ducks, potentially affecting their subsequent growth and development. As computer technology advances and artificial intelligence matures, contactless measurement methods leveraging computer vision and AI are increasingly expected to replace traditional techniques. These innovations offer promising opportunities for reducing labor demands, minimizing measurement errors, and supporting stress-free data collection, paving the way for precision farming applications.

Researchers around the world have extensively studied the application of computer technology in livestock farming, particularly in poultry farming\citep{9309300}. Early research often employed traditional machine learning algorithms. As early as 2009, \cite{4913289} used an elliptical model to approximate the shape of poultry, employing particle filtering algorithms to predict the location and movement of poultry, and tracking algorithms that included predicting positional changes, calculating likelihood, and resampling particles to accurately estimate the state of the poultry. In 2012, \cite{hepworth2012broiler} used Support Vector Machines (SVM) to identify features related to hock burn in chickens, achieving an accuracy of 78\% . Furthermore, watershed algorithms\citep{mortensen2016weight}, K-Means clustering\citep{zhuang2018development}, and Mean Shift\citep{mansor2013approach} clustering are also commonly used for background segmentation in poultry images.


With the continuous development of deep learning, neural networks are increasingly used for predicting poultry characteristics. In \citep{MORTENSEN2016319}, researchers used a depth camera to capture the depth images of broilers and constructed an approximate 3D model. Based on this, they employed geometric methods to extract both 2D and 3D features of the broilers. The features were then used to train a Bayesian Artificial Neural Network (Bayesian ANN) to develop a model predicting the weight of the broilers, with a mean relative error (MRE) and relative standard deviation (RSD) of 7.8\% and 6.6\%, respectively. Although this approach innovatively incorporated a Bayesian ANN, the feature extraction stage still relied on geometric methods, making the process relatively complex and limiting its generalizability and fault tolerance. Convolutional neural networks (CNNs) have unique advantages in image feature extraction. \cite{ZHUANG2019106} improved the CNN-based object detection model, Single Shot MultiBox Detector (SSD)\citep{10.1007/978-3-319-46448-0_2}, and proposed the Improved Feature Fusion Single Shot MultiBox Detector (IFSSD) for real-time monitoring of broiler health, achieving an average precision mean (mAP) of 99.7\%.


The Transformer, originally a neural network structure designed for natural language processing (NLP) incorporating attention mechanisms \citep{NIPS2017_3f5ee243}, reached a climax in the field of computer vision with the introduction of the Vision Transformer (ViT) by \cite{dosovitskiy2021image}. \cite{10.1145/3577148.3577155} used the improved ViT-based classification model, Transformer Fine Grained model (TransFG), to predict the reproductive performance of roosters through images of their combs. The results surpassed other CNN models by 1.4\% in mAP. Although ViT exhibits superior global perception capabilities, it lacks training efficiency and local perception capabilities compared to CNNs, particularly performing poorer on smaller datasets. Therefore, combining the strengths of Transformers and CNNs to enhance model performance becomes particularly crucial \citep{diagnostics11081384}. \cite{HE2023102459} combined Transformer and CNN to propose the Residual Transformer Fine Grained (ResTFG) for fine-grained classification of seven species of chicken Eimeria on microscopic images, achieving superior precision and inference speed compared to other models.


In recent years, significant progress has been made in the prediction of body dimensions for large livestock using computer vision and sensor technologies \citep{s24051504}. For example, \cite{LINAZHANG201833} applied image processing techniques combined with SLIC superpixels and fuzzy clustering algorithms of C-means to perform foreground segmentation, centerline extraction and automatic measurement point identification, enabling the estimation of body dimensions for sheep. Similarly, \cite{DU2022107059} employed a 2D-3D fusion approach, utilizing deep learning models to detect key points in RGB images, which were then projected onto the surface of point clouds. By integrating interpolation and pose normalization techniques, their method achieved automatic measurement of multiple body dimension parameters for cattle and pigs, with MAPEs reduced to below 10\%. Moreover, in \citep{HAO2023107560}, researchers improved the PointNet++ point cloud segmentation model by subdividing point clouds into local regions such as the head, ears, and torso. This refinement enhanced the accuracy of key point localization for measurements, and with additional geometric processing algorithms, the relative errors in multiple body dimension parameters for pigs were significantly reduced.

Despite these advancements, research on body dimension prediction remains confined to large livestock, and studies addressing the prediction of weight and body dimensions for poultry are virtually non-existent. To bridge this gap, this study proposes a neural network model based entirely on deep learning techniques, utilizing the extraction and fusion of 3D spatial features and 2D image features to predict the weight and body dimensions of ducks. This is the first computer vision-based model aimed at weight and body dimension prediction for poultry. The primary objectives of this study are as follows:
\begin{enumerate}
    \item To propose a comprehensive hardware-software integrated multidimensional and multi-view visual data acquisition scheme for ducks, and it is utilized to collect a dataset of duck visual data along with their corresponding body dimensions and weight.
    \item To propose a method combining PointNet++ to identify key points in the point cloud and compute the 3D geometric features of the duck.
    \item To propose a deep learning model combining 2D convolutional features and 3D geometric features to predict the body dimensions and weight of the duck.
    \item To evaluate the performance and effectiveness of the model and discuss potential avenues for future improvements.
\end{enumerate}



\section{Data Collection}
\label{Data Collection}

\subsection{Dataset Description}
A suitable dataset is crucial for deep learning applications. For the prediction of duck body dimensions, it is essential to have visual information of the ducks, along with corresponding ground truth measurements as annotations. However, to date, there is no publicly available dataset in this field that meets these requirements. Therefore, we collected the dataset by visiting Hunan Linwu Shunhua Duck Industrial Development Co., Ltd., located in Linwu County, Hunan Province, China. We obtained side-view depth images, RGB images, and top-view RGB images of 1,023 ducks, and measured their corresponding morphometric parameters as shown in \autoref{Table1} To enhance the generalization capability of the model, multiple sets of visual information were typically collected for each duck in different postures and states. In our research, 5,238 sets of visual information of 1,023 Linwu ducks were collected.

\begin{table*}
\centering
\begin{tabular}{l c c} 
 \hline
 Parameter & Unit & Definition \\ 
 \hline
 Weight          & g  & The overall weight of the duck. \\ 
 Body Diagonal Length     & cm & The diagonal length from the tip of the beak to the tail. \\ 
 Neck Length     & cm & The length of the duck's neck, from the base to the head. \\ 
 Semi-Diving Length & cm & The depth the duck's body enters the water while diving. \\ 
 Keel Length     & cm & The length of the duck's keel bone, influencing chest development. \\ 
 Chest Width     & cm & The width of the duck's chest, indicating chest development. \\ 
 Chest Depth     & cm & The vertical distance from the back to the abdomen, reflecting chest depth. \\ 
 Tibia Length    & cm & The length of the duck's tibia, associated with its mobility. \\ 
 \hline
\end{tabular}
\caption{Duck morphometric parameters with corresponding units and definitions.}
\label{Table1}
\end{table*}

\subsection{Collection Method}


During the visual data acquisition process, we utilized the Intel RealSense D415 depth camera, which is capable of simultaneously capturing RGB and depth images beyond a distance of 0.4 meters. Additionally, we employed the Logitech C270 camera by Logitech as an RGB camera to capture images from another angle. Several high-powered supplementary lights were used to enhance brightness, particularly in poorly lit environments.

In the actual measurement of the duck’s morphometric parameters, the weight was measured using an electronic weighing scale. The body diagonal length, neck length, semi-diving length, and keel length were measured using a measuring tape, while the chest width, chest depth, and tibia length were measured using an electronic caliper. These measurements were manually recorded as annotations.

\begin{figure}[h]
    \centering
    \includegraphics[width=\linewidth]{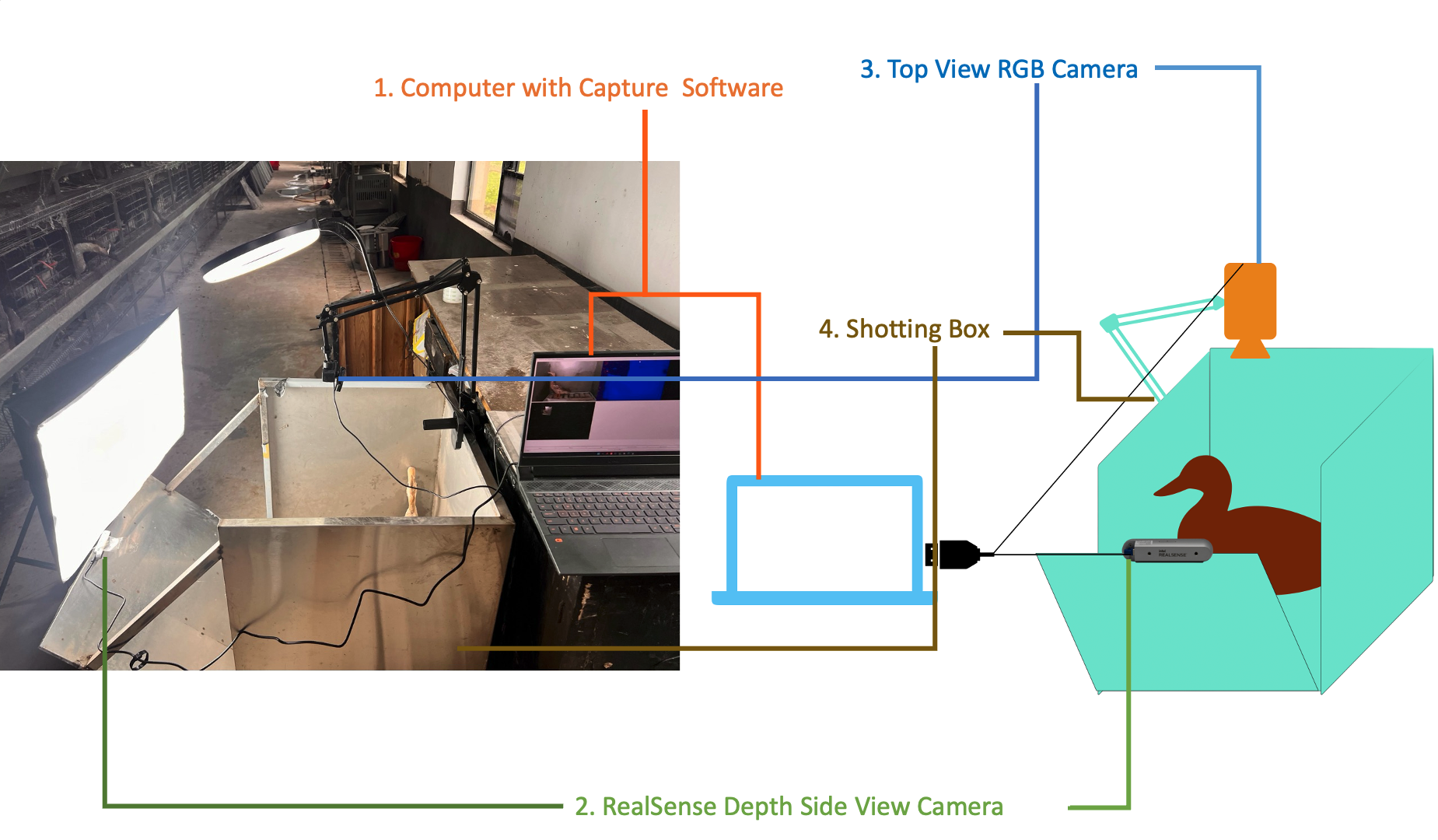} 
    \caption{Collection devices and their layout.}
    \label{fig:0}
\end{figure}

Ducks, particularly the older Linwu ducks, are inherently lively and exhibit strong reactions to human contact, making it extremely challenging to capture visual information. Therefore, during photography, a relatively enclosed space is required to keep the ducks calm. We designed a relatively enclosed shooting box, as shown in \autoref{fig:0}, and arranged the imaging equipment on the top and sides of the box. Given that the depth camera has specific distance requirements for capturing objects, and that too close a distance prevents the camera from capturing the full image of the duck, it is necessary to maintain a certain distance between the camera and the subject. However, an excessive shooting distance would create additional space, giving the duck more room to move, and making it easier for the duck to escape if the side with the camera is left open. To address this, we designed the shooting side as a 45-degree inclined plane and positioned the RealSense depth camera above the slope. This design not only creates an enclosed environment for the duck but also provides adequate space for capturing images through the extended distance created by the inclined plane.

\begin{figure}[h]
    \centering
    \includegraphics[width=\linewidth]{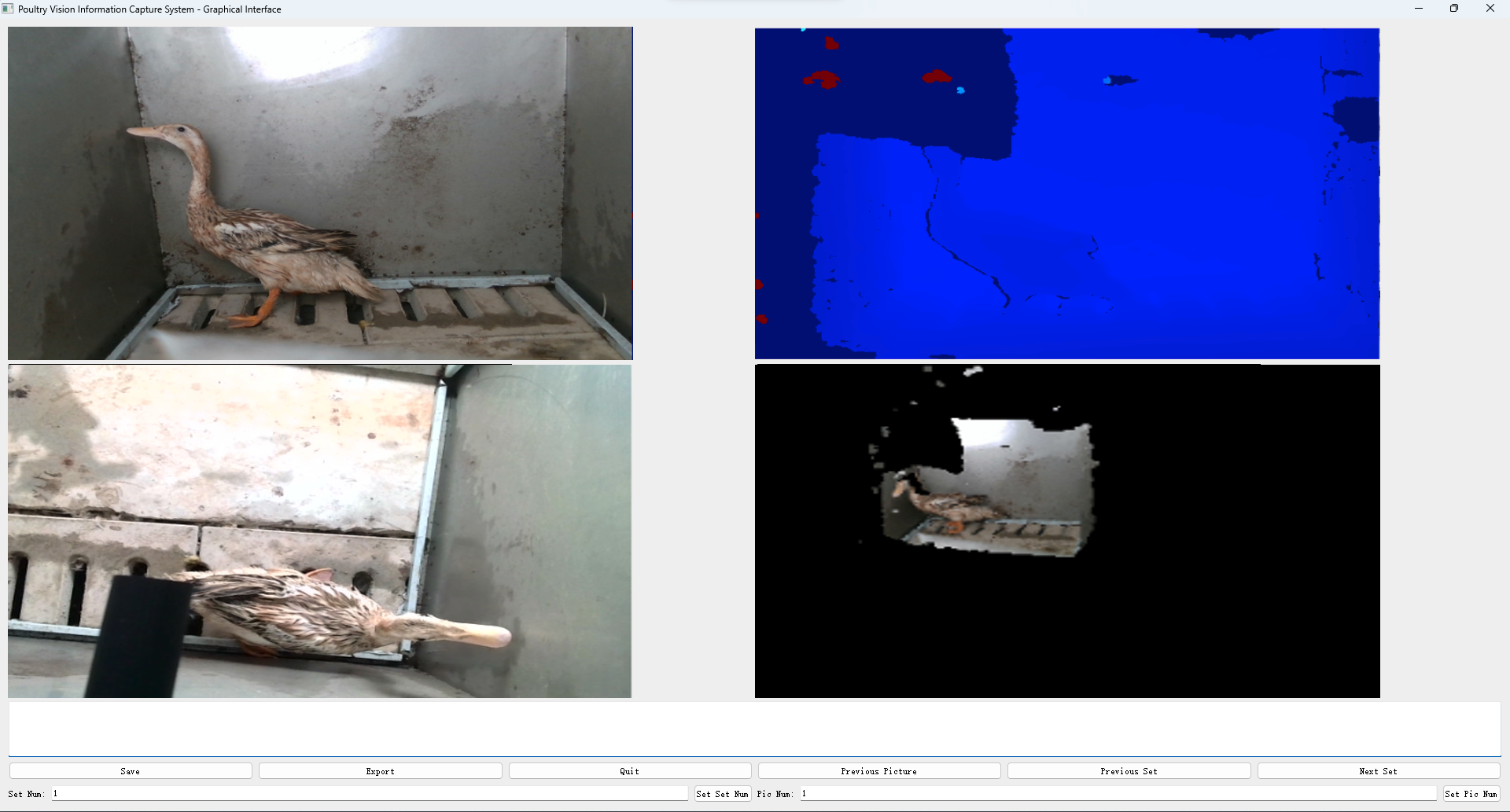} 
    \caption{Capture software.}
    \label{fig:0.5}
\end{figure}

As illustrated in \autoref{fig:0.5}, we developed a software system for capturing visual information. This software is capable of simultaneously capturing data from multiple camera angles and can automatically align the RealSense depth images with the RGB images during shooting. Additionally, it generates and saves point cloud data. The software also features automatic numbering, export, and rollback functions for operational convenience.

\section{Data Processing}
\label{Data Processing}

\subsection{Features Extraction}
\label{Features Extraction}

Since the body dimensions of ducks are precise numerical values, the effectiveness of allowing neural networks to directly extract features from images may be suboptimal, particularly when the dataset is small. The specific differences will be compared and discussed in detail in \autoref{Results and Discussions}. Therefore, it is necessary to incorporate additional feature extraction steps to obtain more direct information.


A point cloud is a data structure that represents the shape and spatial information of an object by capturing thousands or millions of three-dimensional coordinates of points on the object’s surface. It is widely used in 3D modeling, computer vision, and environmental perception \citep{5980567}. The point cloud images collected during our data acquisition process can provide more intuitive characteristic information of the ducks. In \citep{s19225046}, the researchers developed the D Point Cloud model to directly predict key points associated with cattle body dimensions, which were subsequently used to calculate the body dimension data of cattle. However, for poultry, the presence of thick feathers and greater morphological flexibility make it nearly impossible to directly calculate body dimensions from the key points of ducks. We aim to identify key points in the point cloud that are related to the body dimensions of ducks and use the geometric relationships between these points as features to provide additional information for subsequent body dimension predictions.

First, we applied a series of steps to denoise and filter the point cloud data. Initially, we used a statistical filtering method to identify and remove noise points by calculating the average distance of each point to its 20 nearest neighbors and comparing it to the global average distance. We set a standard deviation multiplier threshold of 2.0, meaning that any point whose average distance exceeds twice the global average distance is considered noise and subsequently removed. Following this denoising process, we performed a clustering analysis on the point cloud data and further filtered the clusters by retaining only those with a point count of 9,000 or more. This approach not only effectively removed isolated and anomalous points but also ensured that the remaining point cloud clusters were sufficiently dense and representative, thereby enhancing the accuracy and reliability of subsequent processing steps.

In the denoising process, the average distance \( \bar{d}_i \) between each point \( p_i \) and its \( k \) nearest neighbors is calculated:

\begin{equation}
\bar{d}_i = \frac{1}{k} \sum_{j=1}^{k} \| p_i - p_{ij} \|
\end{equation}

where \( p_{ij} \) denotes the \( j \)-th nearest neighbor of point \( p_i \), and \( k = 20 \) is the number of nearest neighbors.

This average distance is then compared with the global average distance \( \bar{d}_{\text{global}} \). A standard deviation multiplier threshold \( \sigma_{\text{threshold}} = 2.0 \) is set. If \( \bar{d}_i \) satisfies the following condition, the point is considered noise and removed:

\begin{equation}
\bar{d}_i > \bar{d}_{\text{global}} + 2.0 \times \sigma_{\text{global}}
\end{equation}

where \( \sigma_{\text{global}} \) represents the standard deviation of the global distances.

\begin{figure}[h]
    \centering
    \includegraphics[width=0.8\linewidth]{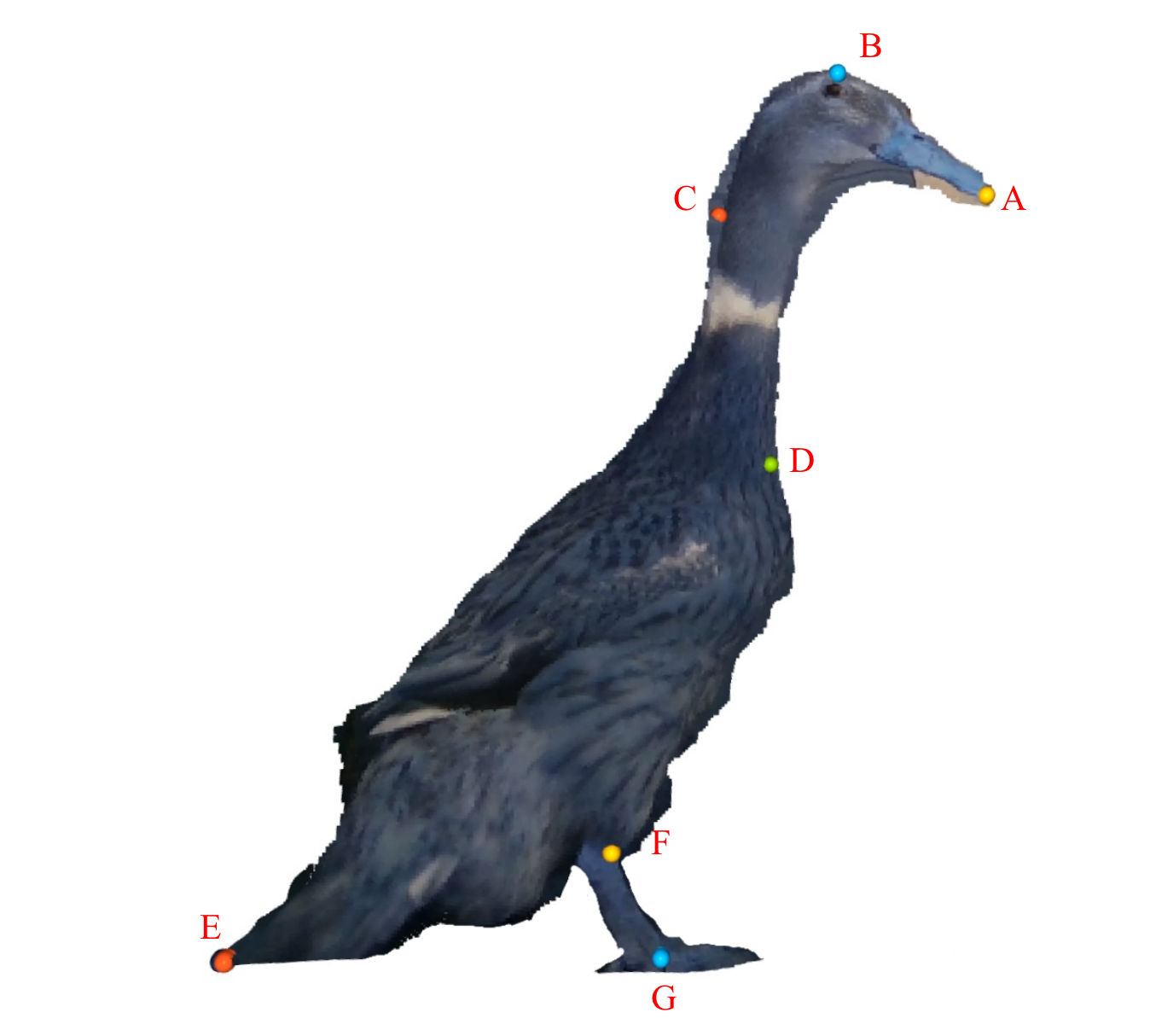} 
    \caption{Seven annotated feature points on the duck's point cloud used for geometric feature extraction, including key anatomical landmarks such as the beak tip, head peak, neck curve, and tail tip}
    \label{fig:1}
\end{figure}

Subsequently, as shown in \autoref{fig:1}, we selected seven relevant points as feature points.
\begin{itemize}
    \item \textbf{Point A}: Located at the foremost tip of the duck's beak.
    \item \textbf{Point B}: At the highest point of the duck's head.
    \item \textbf{Point C}: At the most prominent point where the duck's neck curves towards the tail.
    \item \textbf{Point D}: At the junction between the duck's neck and chest.
    \item \textbf{Point E}: Located at the very end of the duck's tail.
    \item \textbf{Point F}: At the top of the duck's foot.
    \item \textbf{Point G}: At the bottom of the duck's foot.
\end{itemize}

Although not all of these points are necessarily meaningful from the perspective of avian science, they are distinctive and easily learnable features from a computational viewpoint. We developed a simple software tool for annotating feature points in point clouds and manually annotated the above key points in 150 point cloud images.

This translation is crafted to ensure clarity and precision, adhering to academic standards while effectively communicating the process and rationale behind the selection and annotation of the feature points.


Given that each processed point cloud image still contains over 200,000 points, it is essential to reduce the complexity to a manageable scale for deep learning models. We employed Farthest Point Sampling (FPS)\citep{NIPS2017_d8bf84be} to uniformly downsample all point clouds to a consistent size of 8192 points per sample. FPS ensures spatial uniformity by iteratively selecting points that maximize the minimum distance to previously selected points, thus preserving critical geometric details. Formally, the FPS algorithm selects a subset of points \(S\) from the original set of points \(P\) using the following iterative criterion:
\begin{equation}
p_{new} = \arg\max_{p \in P \setminus S}\left(\min_{q \in S}\|p - q\|\right)
\end{equation}
This method maintains an optimal spatial distribution, critical for capturing anatomical features accurately.

As shown in \autoref{fig:3} (b), the processed 8192-point clouds were then fed into a modified PointNet++ model \citep{NIPS2017_d8bf84be}, a hierarchical neural network recognized for effectively capturing local-to-global point cloud features through Set Abstraction (SA) layers. The original PointNet++ architecture, intended for classification and segmentation tasks, was adjusted to perform precise keypoint regression by altering the final fully connected layers.

Specifically, we retained the hierarchical feature extraction approach of PointNet++, which consists of sequential SA layers that progressively aggregate local and global spatial information. To perform regression tasks, the output classification or segmentation heads of PointNet++ were replaced with fully connected regression layers. Mathematically, the modified fully connected layers transform the global feature vector \(\mathbf{F}_{global}\) (extracted from the last SA layer) through two sequential non-linear mappings followed by a linear mapping, which directly predicts the three-dimensional coordinates of the seven keypoints. Formally, this transformation is defined as:
\begin{equation}
\mathbf{Y} = \mathbf{W}_3\left(\text{ReLU}\left(\mathbf{W}_2\left(\text{ReLU}(\mathbf{W}_1 \mathbf{F}_{global} + \mathbf{b}_1)\right)+\mathbf{b}_2\right)\right) + \mathbf{b}_3
\end{equation}
where \(\mathbf{W}_i\) and \(\mathbf{b}_i\) are weight matrices and bias vectors of the fully connected layers. The resulting output tensor dimension is \(7 \times 3 = 21\), corresponding exactly to the coordinates of the seven predefined feature points. After training for 40 epochs, our modified PointNet++ achieved robust performance, evidenced by a mean squared error (MSE) of only 0.0009 on the test dataset.

After identifying the seven feature points for each point cloud sample, we extracted ten key geometric features from the seven points to describe their spatial relationships. Specifically, we calculated the following:
\begin{enumerate}
    \item \textbf{Distances between points:}
    \begin{itemize}
        \item Distance between points A and B
        \item Distance between points B and C
        \item Distance between points C and D
        \item Distance between points D and E
        \item Distance between points E and F
        \item Distance between points F and G
    \end{itemize}
    
    \item \textbf{Angles formed by points:}
    \begin{itemize}
        \item Angle between points A, B, and C
        \item Angle between points B, C, and D
        \item Angle between points C, D, and E
        \item Angle between points D, E, and F
    \end{itemize}
\end{enumerate}

These feature values comprehensively reflect the relative positions and arrangements of the points, providing critical geometric information for our analysis.

\subsection{Data Prepossessing}




In the dataset, many visual data were of low quality due to factors such as lighting conditions, accidental occlusions, or duck movements during capture, rendering them unsuitable for training. Consequently, we filtered out a portion of the images that did not meet the required standards. Ultimately, a total of 4,822 sets of image data were used for training. 

In our study, during the preprocessing of depth data, we retained only the depth information within the range of 40cm to 1200cm, as the subject of our images—ducks—typically remains within this distance. Depth values outside this range were considered noise and thus filtered out. Subsequently, the retained depth data were normalized to a standard grayscale range suitable for image representation. This preprocessing step effectively transformed the multidimensional depth data into a visual format that is easier to analyze and interpret.

The background of the images contains unnecessary content and thus needs to be removed. We employed U-Net \citep{ronneberger2015u}, a CNN architecture designed primarily for biomedical image segmentation, characterized by its U-shaped structure that enables precise localization and classification by combining high-resolution features from the contracting path with upsampled outputs from the expanding path, to segment the background in images captured from side and top views. In our experiments, excellent segmentation results were achieved by annotating only 50 sets of images. Since the depth images were already aligned with the side-view RGB images during acquisition, the depth data can be directly processed for background removal using the corresponding segmentation of the RGB images.
\begin{figure}[h]
    \centering
    \includegraphics[width=\linewidth]{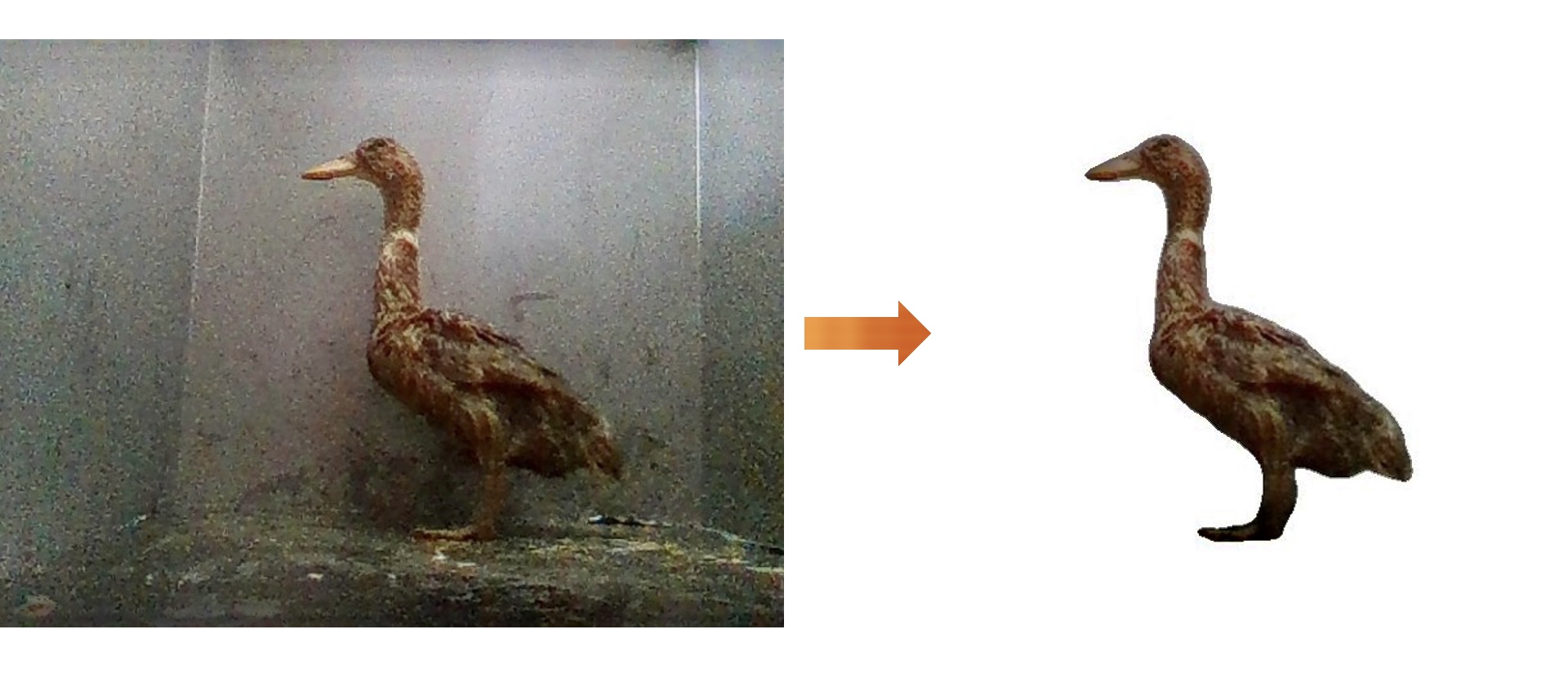} 
    \caption{Segmentation results of the duck's side-view.}
    \label{fig:2}
\end{figure}

For the duck’s body weight and body dimension data, we applied Min-Max normalization to each column, ensuring more stable and reliable training in the subsequent stages.

\section{Method}
\label{Model Architecture}

\begin{figure*}[h]
    \centering
    \includegraphics[width=\linewidth]{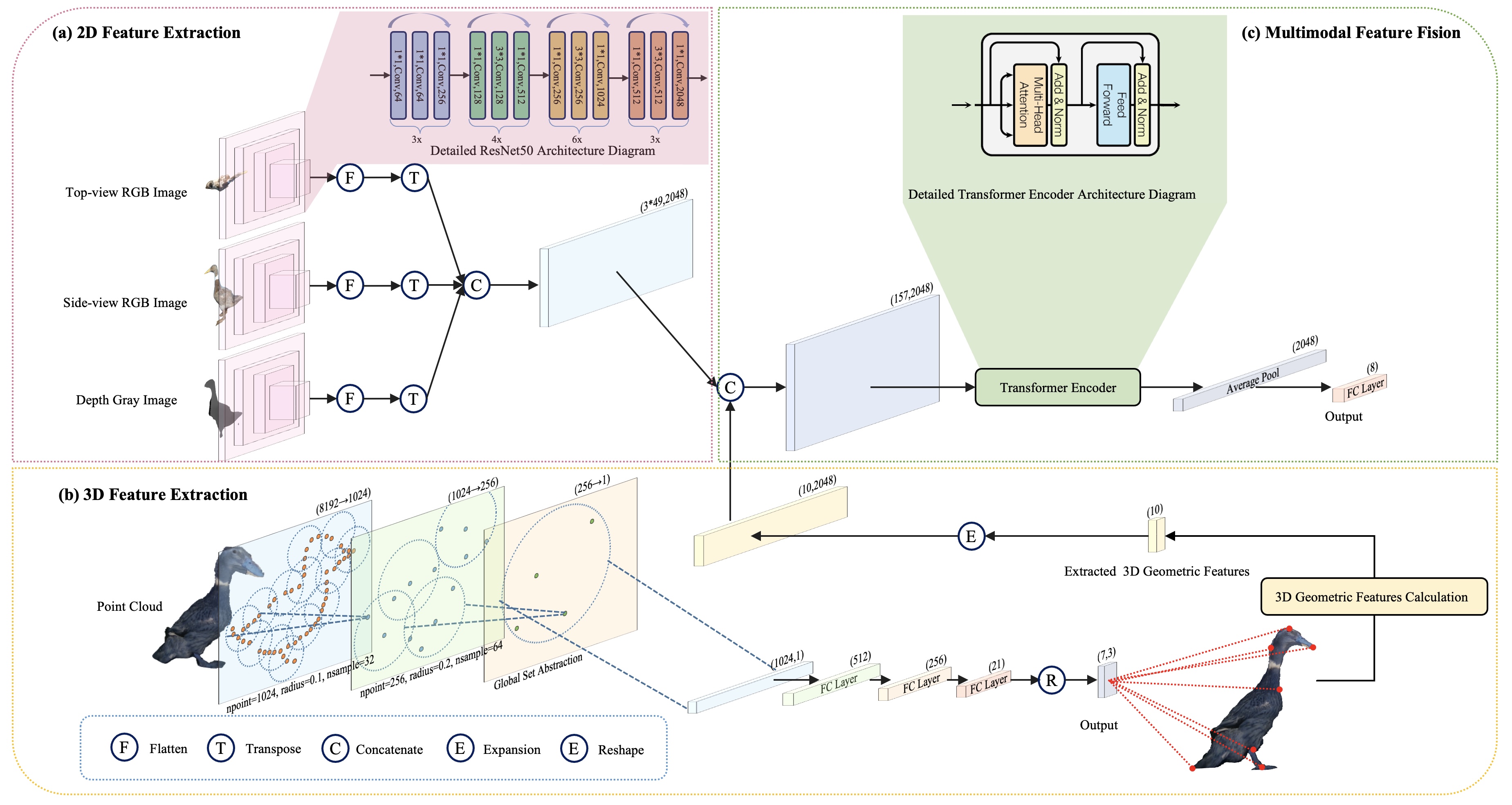} 
    \caption{Architecture of the proposed multimodal model for predicting duck body dimensions and weight. (a) 2D feature extraction: Three independent ResNet50 models extract visual features from top-view RGB, side-view RGB, and depth images, respectively. (b) 3D feature extraction: A modified PointNet++ model performs hierarchical feature extraction from the duck point cloud through three Set Abstraction layers. The final global features are then processed by fully connected layers to predict seven anatomical keypoints in 3D coordinates. (c) Multimodal feature fusion: Image features and 3D geometric features are integrated and refined through a Transformer encoder, enabling accurate prediction by capturing global feature dependencies across different data modalities.}
    \label{fig:3}
\end{figure*}

As shown in \autoref{fig:3}, our proposed model consists of three main components: a Triple ResNet50 embedding module for feature extraction from images, a Transformer encoder module for capturing complex relationships among features, and a regression layer for predicting the duck body dimensions and weight.

To extract rich visual features from the images, we utilize three independent ResNet50 networks \citep{he2016deep}, each pre-trained on ImageNet. Each ResNet50 network processes one of the three input images: the top-view RGB image \( I_{\text{top}} \), the side-view RGB image \( I_{\text{side}} \), and the side-view depth grayscale image \( I_{\text{depth}} \). The final classification layers of ResNet50 are removed, retaining the convolutional layers up to the penultimate layer to serve as feature extractors.

For each image \( I_i \), the corresponding ResNet50 network outputs a feature map \( F_i \in \mathbb{R}^{C \times H \times W} \), where \( C \) is the number of channels, and \( H \) and \( W \) are the spatial dimensions. These feature maps are then reshaped into sequences of feature vectors suitable for the Transformer encoder. Specifically, each feature map \( F_i \) is flattened along the spatial dimensions and transposed to form a sequence \( X_i \in \mathbb{R}^{N \times D} \), where \( N = H \times W \) is the sequence length and \( D = C \) is the feature dimension:
\begin{equation}
X_i = \text{reshape}(F_i) \in \mathbb{R}^{N \times D}
\end{equation}

After obtaining the feature sequences \( X_i \) from the three ResNet50 models, we concatenate them along the sequence dimension to form a combined visual feature sequence:
\begin{equation}
X_{\text{img}} = [X_{\text{top}}, X_{\text{side}}, X_{\text{depth}}] \in \mathbb{R}^{3N \times D}
\end{equation}
In addition to the image-based features, we incorporate ten geometric features \( G \in \mathbb{R}^{10} \) extracted from the point cloud data. These features capture specific spatial relationships between key points on the duck's body that are not directly represented in the image data. To integrate these geometric features with the visual features, we expand \( G \) along the sequence dimension to match the length of \( X_{\text{img}} \):
\begin{equation}
G' = \text{repeat}(G, 3N) \in \mathbb{R}^{3N \times 10}
\end{equation}
We then concatenate \( G' \) with \( X_{\text{img}} \) along the feature dimension to form the final input sequence \( X \):
\begin{equation}
X = [X_{\text{img}}, G'] \in \mathbb{R}^{3N \times (D + 10)}
\end{equation}
This combined feature sequence \( X \) contains both visual and geometric information, enabling the model to capture comprehensive representations of the duck's body from different perspectives.

The integrated feature sequence \( X \) is input into a Transformer encoder module to capture complex spatial relationships and dependencies among the features. The Transformer encoder consists of multiple layers, each comprising a multi-head self-attention mechanism and a position-wise feed-forward network, as described in \citep{NIPS2017_3f5ee243}. The self-attention mechanism allows the model to focus on different parts of the sequence to effectively capture global dependencies. Residual connections and layer normalization are applied around the attention and feed-forward sublayers to facilitate training stability.

The Transformer encoder outputs a refined sequence of feature representations \( Z \in \mathbb{R}^{3N \times (D + 10)} \) that capture the complex interactions among the visual and geometric features.

To aggregate the sequence of feature representations into a fixed-length vector suitable for regression, we apply average pooling over the sequence dimension:
\begin{equation}
z = \frac{1}{3N} \sum_{i=1}^{3N} Z_i \in \mathbb{R}^{D + 10}
\end{equation}
Finally, a fully connected layer maps the pooled feature vector to the target outputs, which are the eight body dimension measurements:
\begin{equation}
\hat{y} = z W + b \in \mathbb{R}^{8}
\end{equation}
where \( W \in \mathbb{R}^{(D + 10) \times 8} \) and \( b \in \mathbb{R}^{8} \) are the weights and biases of the regression layer.

By integrating CNNs and Transformer architectures, our model effectively captures both local visual features and global geometric relationships, enabling accurate estimation of duck body dimensions from multi-modal data sources.

\section{Results and Discussions}
\label{Results and Discussions}

\subsection{Results}





We utilized PyTorch 2.1.0 with CUDA 12.1 in an environment equipped with two Nvidia Tesla T4 GPUs for model training. The dataset was split into training, validation, and testing sets in a ratio of 8:1:1. After approximately 20 epochs of training, the model gradually converged.

\begin{figure*}[htbp]
    \centering
    \includegraphics[width=0.97\linewidth]{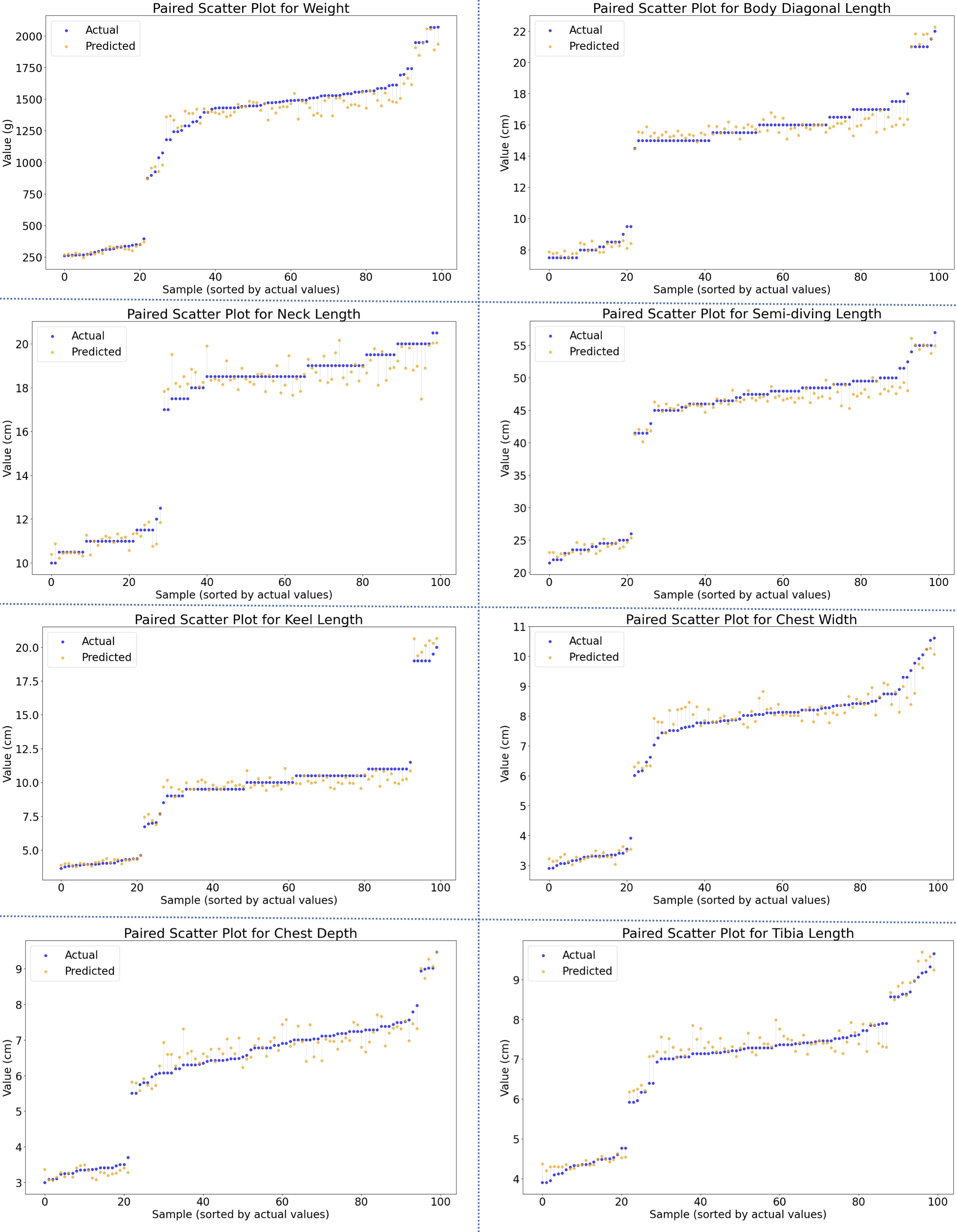} 
    \caption{Paired scatter plots for eight parameters: Weight, Body Diagonal Length, Neck Length, Semi-diving Length, Keel Length, Chest Width, Chest Depth, and Tibia Length.}
    \label{fig:4}
\end{figure*}

The best-performing model was selected for weight and body size prediction, achieving the metrics results presented in \autoref{Table2}, with an overall MAPE of 6.33\% and R\textsuperscript{2} of 0.953. \autoref{fig:4} presents the comparison between the actual and predicted values of 100 samples from the test set, sorted in ascending order based on the actual values. The parameters include weight, body diagonal length, neck length, semi-diving length, keel length, chest width, chest depth, and tibia length.

\begin{table*}[htbp]
\centering
\begin{tabular}{lccccc}
\hline
\textbf{Morphometric Parameters} & \textbf{R\textsuperscript{2}} $\uparrow$ & \textbf{MAPE (\%)} $\downarrow$ & \textbf{RMSE} $\downarrow$ & \textbf{MAE} $\downarrow$ \\
\hline

Body Diagonal Length (cm)& 0.968  & 5.17  & 0.813  & 0.651   \\
Neck Length (cm)         & 0.927  & 5.89  & 1.120  & 0.804   \\
Semi-diving Length (cm)   & 0.966  & 4.32  & 2.298  & 1.687   \\
Keel Length (cm)         & 0.973  & 6.77  & 0.773  & 0.577   \\
Chest Width (cm)         & 0.952  & 6.34  & 0.563  & 0.387   \\
Chest Depth (cm)         & 0.931  & 6.92  & 0.517  & 0.385   \\
Tibia Length (cm)        & 0.955  & 4.74  & 0.381  & 0.297   \\
Overall (Body Dimensions)& 0.953  & 5.73  & 0.924  & 0.684 \\
\hline
Weight (g)               & 0.952  & 10.49 & 135.0  & 96.63   \\

\hline
Overall (All)                  & 0.953  & 6.33  & 17.68  & 12.68   \\
\hline
\end{tabular}
\caption{Performance metrics for ducks' weight and body dimensions in test set.}
\label{Table2}
\end{table*}

Experimental results demonstrate that our model achieved excellent performance in predicting both the weight and various body dimension metrics of ducks, with results highly comparable to manual measurements. This is especially notable considering potential errors introduced during dataset collection, such as weight variations due to the ducks' fed or fasting state and the condition of their feathers (e.g., wet or dry), as well as errors arising from manual measurements of body dimensions.


In our experiments, we systematically evaluated several model architectures, including CNN-based models (VGG16, VGG19\citep{2014arXiv1409.1556S}, ResNet34, ResNet50, and ResNet101\citep{he2016deep}) and Transformer-based visual models (ViT-B/16, ViT-L/16\citep{dosovitskiy2021image}, and Swin-T\citep{DBLP:journals/corr/abs-2103-14030}), as summarized in \autoref{Table3}. We examined model performance variations when incorporating or excluding key geometric feature extraction and Transformer encoders.

The experimental results shown in \autoref{Table3} clearly indicate that the integration of CNN-based backbones, particularly ResNet50, along with feature points extraction and Transformer encoder, achieves optimal performance (with the highest $R^2$ of 0.953 and the lowest errors in terms of MAPE, RMSE, and MAE). Thus, all three modules (CNN backbone, feature point extraction, and Transformer encoder) significantly contribute to the model’s accuracy. Additionally, visual Transformer-based models (ViT and Swin-T) demonstrated relatively lower performance compared to ResNet-based architectures, indicating that Transformer-based backbones alone are less effective for capturing essential local spatial features with the current dataset size and task complexity.

\begin{table*}[htbp]
\centering

\begin{tabular}{lccccc}
\hline
\textbf{Model Architecture} & \textbf{R\textsuperscript{2}} $\uparrow$ & \textbf{MAPE (\%)} $\downarrow$ & \textbf{RMSE} $\downarrow$ & \textbf{MAE} $\downarrow$ \\
\hline

VGG16 + Feature Points Extraction + Transformer Encoder        & 0.929 & 8.88  & 19.69  & 15.38  \\
VGG19 + Feature Points Extraction + Transformer Encoder         & 0.931 & 7.97  & 18.11  & 14.99  \\
ViT-L/16 + Feature Points Extraction + Transformer Encoder      & 0.832 & 15.27 & 37.88  & 31.83  \\
ViT-B/16 + Feature Points Extraction + Transformer Encoder      & 0.844 & 10.47 & 28.16  & 20.01 \\
Swin-T + Feature Points Extraction + Transformer Encoder      & 0.940 & 7.10 & 18.12  & 12.76 \\
ResNet34 + Feature Points Extraction + Transformer Encoder      & 0.888 & 10.86 & 20.51  & 16.39  \\
ResNet101 + Feature Points Extraction + Transformer Encoder     & 0.935 & 7.46  & 18.22  & 13.39  \\
ResNet50 + Transformer Encoder                                   & 0.903 & 7.90  & 18.95  & 12.83  \\
ResNet50 Only                                           & 0.813 & 17.08 & 40.30  & 35.11  \\
\textbf{ResNet50 + Feature Points Extraction + Transformer Encoder}      & \textbf{0.953} & \textbf{6.33}  & \textbf{17.68}  & \textbf{12.68}  \\
\hline
\end{tabular}
\caption{Performance metrics for different model architectures.}
\label{Table3}
\end{table*}

\subsection{Discussions}

The predictive results of this study are primarily derived from deep learning-based training and inference, with minimal reliance on explicit prior knowledge. Compared to existing studies on the prediction of weight and body dimensions of large livestock \citep{agriculture14020306}, our approach emphasizes a data-driven methodology, representing a significant innovation. While previous methods often integrated explicit prior knowledge into neural networks or machine learning algorithms for predicting livestock weight and body dimensions, our proposed model focuses on automatically learning relevant features from multimodal data, making it both novel and effective.

\begin{figure}[htbp]
    \centering
    \includegraphics[width=\linewidth]{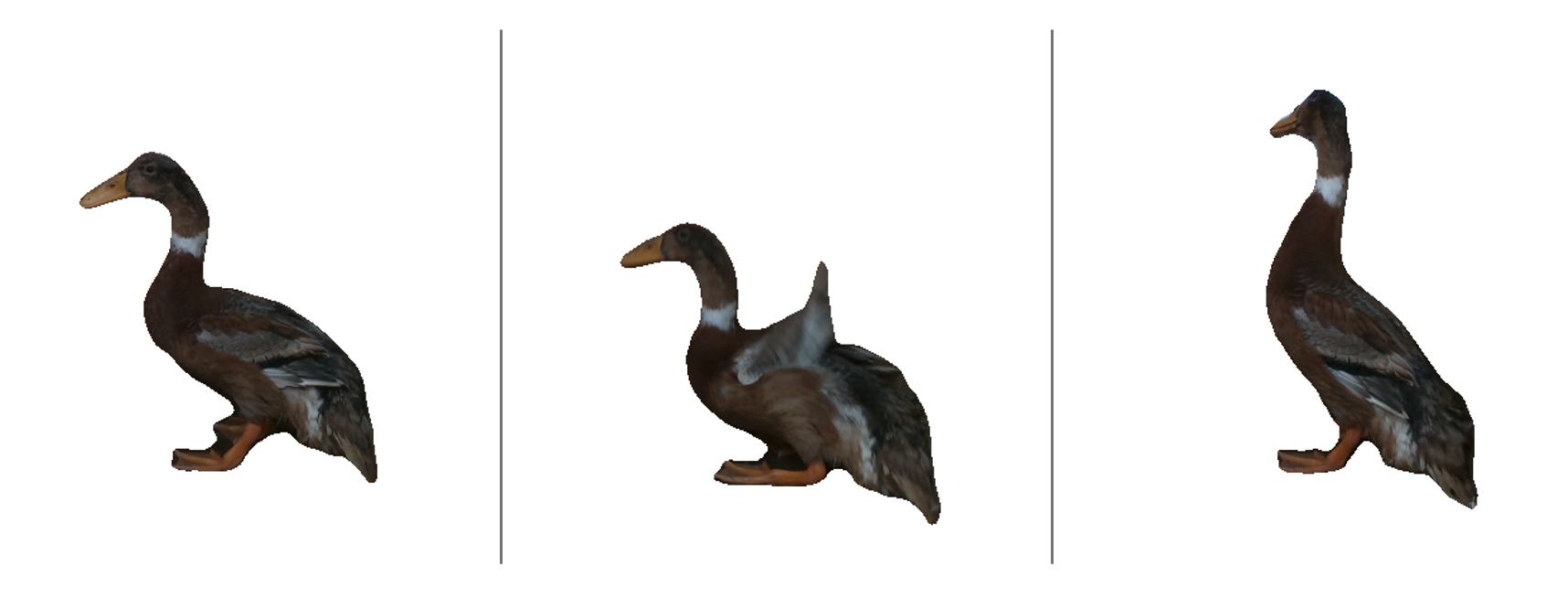} 
    \caption{Three side-view RGB images (segmented) of the same duck in different poses}
    \label{fig:5}
\end{figure}

\begin{figure}[H]
    \centering
    \includegraphics[width=\linewidth]{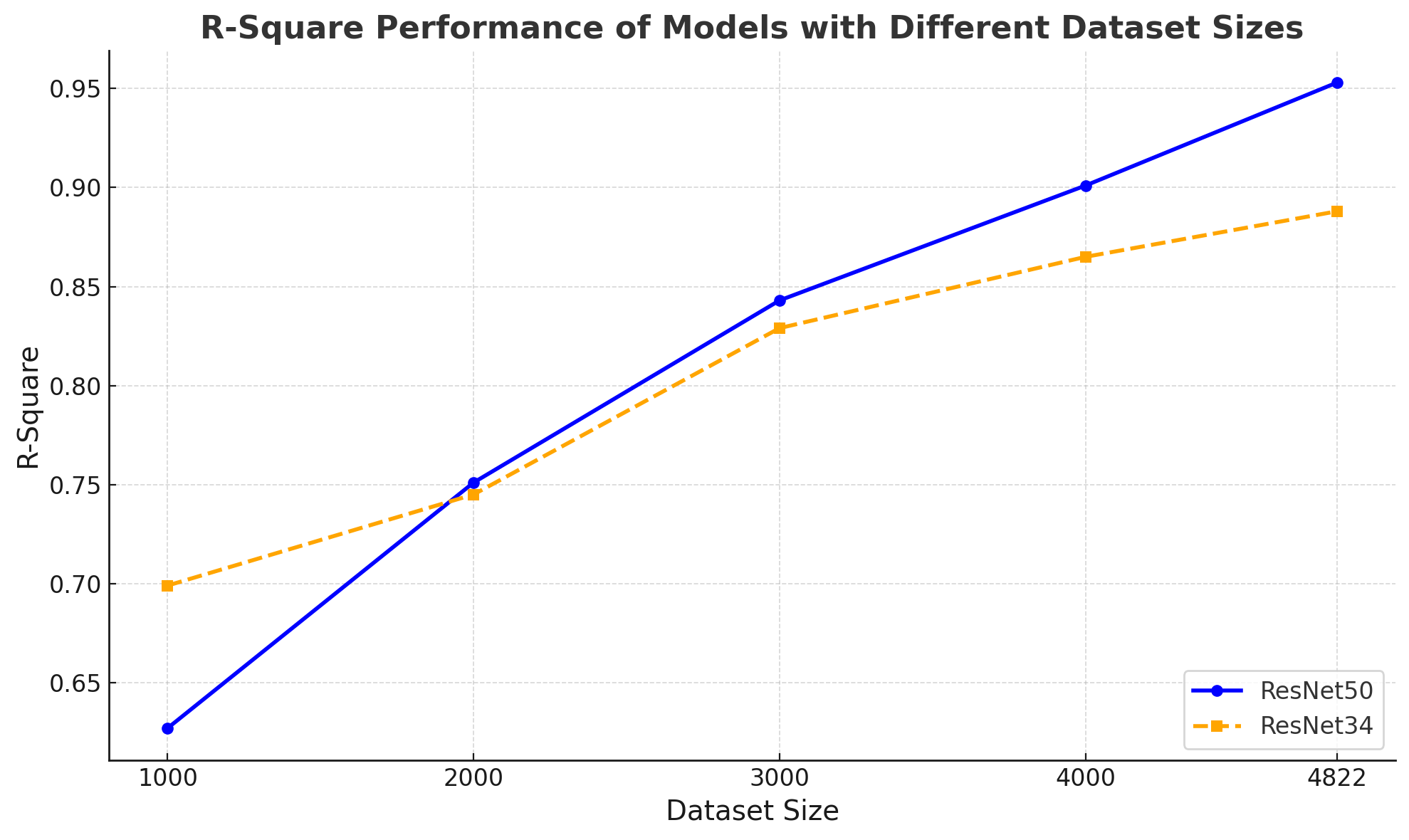} 
    \caption{Comparison of \( R^2 \) performance between ResNet50 and ResNet34 as 2D image feature extractors under varying dataset sizes.}
    \label{fig:6}
\end{figure}

We have identified two critical factors contributing to the success of deep learning methods in predicting poultry weight and body dimensions in this study. First, as observed in \citep{HAO2023107560} for the prediction of pig body dimensions, non-standard postures of animals can influence training effectiveness and prediction accuracy. This is particularly pronounced in poultry due to their greater range of motion, more active nature, and difficulty maintaining standard postures under bright lights or stress in shooting environments. To address this, we collected multiple sets of visual data from different postures of the same duck, as illustrated in \autoref{fig:5}. This approach allowed the deep learning model to learn the relationships between various postures of the same individual during training. Experimental results confirmed the effectiveness of this method.
Second, the reduced reliance on explicit prior knowledge necessitates larger datasets to achieve optimal model performance. Initially, the dataset for this study was relatively small, leading to suboptimal prediction outcomes. Through continuous data collection and augmentation, the prediction results improved significantly with the increase in sample size. To illustrate this point, we compared the \( R^2 \) performance between ResNet50 and ResNet34 as 2D image feature extractors across varying dataset sizes, and presented the results in \autoref{fig:6}.

Despite these achievements, our study has several limitations. Due to experimental constraints, we were unable to collect data samples covering all age groups of ducks and instead relied on ducks from specific age groups. This limitation likely contributes to the observed gaps in the actual values in \autoref{fig:4}. Future studies could address this by conducting longitudinal data collection to cover the full life cycle of ducks, resulting in a larger and more diverse dataset. Such an effort is expected to further enhance the model's predictive accuracy and robustness.

Additionally, the key points selected for analysis in \autoref{Features Extraction} were derived based on their potential relevance to phenotypic features that may exhibit linear or non-linear correlations with body dimensions, rather than being directly measurable anatomical landmarks. This approach aims to balance computational feasibility and predictive utility. Nevertheless, further research is needed to explore and validate more systematic and data-driven methods for identifying optimal feature points that can better capture the geometric characteristics associated with duck body dimensions.





\section{Conclusions}
\label{Conclusions}




This study is the first to propose a method for measuring poultry body dimensions based on visual sensors and computer vision technology. By leveraging multimodal fusion of 3D geometric features and 2D image features, it enables non-invasive and contactless estimation of the weight and body dimensions of ducks. The estimations achieved in this study are primarily based on deep learning, demonstrating that with the continuous advancements in deep learning within the field of computer vision, it is increasingly feasible to rely on neural networks to learn relevant features from visual information for predicting animal weight and body dimensions. This significant advancement provides new possibilities for future research in this domain. Subsequent studies could focus on improving the model's ability to generalize across different datasets, optimizing feature fusion strategies, and addressing challenges such as small dataset scenarios through data augmentation or transfer learning approaches, while continuing to expand and refine the dataset to enhance model performance.

Given the shared phenotypic characteristics among various poultry species, the proposed model and measurement method in this study can potentially be transferred to the estimation of weight and body dimensions for other poultry, such as chickens and geese, through the collection of corresponding visual datasets. Further research is required to supplement and refine this application.

Moreover, since the unit cost of poultry is significantly lower than that of large livestock, the poultry industry is more sensitive to cost considerations. Enhancing the affordability of this technology and integrating computer vision techniques into the commercial operations of poultry farming to promote industry development is an area that merits the attention of researchers.

\section*{Acknowledgements}
This work was supported by the Scientific Research Key Project of the Education Department of Hunan Province, China [24A0176], 
National Key Research and Development Program[2021YFD1300404 \& 2022YFD1600902-4], 
and National Natural Science Foundation of China [62402170].

\bibliographystyle{elsarticle-harv} 
\bibliography{main}






\end{document}